\documentclass[english, final]{IEEEtran}
\usepackage[T1]{fontenc}
\usepackage{mathrsfs}
\usepackage{bm}
\usepackage{amsmath}
\usepackage{amsthm}
\usepackage{amssymb}
\usepackage{stmaryrd}
\usepackage{graphicx}
 \PassOptionsToPackage{normalem}{ulem}
 \usepackage{ulem}
 \usepackage{cite}
 \usepackage{comment}
 \usepackage{bm}
\usepackage{physics}
\providecommand{\idx}[1]{\textbf{\textit{Index terms---}} #1}

\usepackage[outdir=./]{epstopdf}

\makeatletter

\theoremstyle{PDAin}

  \theoremstyle{PDAin}
  
  \theoremstyle{PDAin}
  
  \theoremstyle{PDAin}
   
  \theoremstyle{remark}

\theoremstyle{assumption}

\theoremstyle{algorithm}

\usepackage{amsmath}
\usepackage{amssymb}
\usepackage{graphicx,psfrag,cite,subfigure}
\usepackage[table]{xcolor}

\usepackage[acronym]{glossaries}

\newcommand{\xddots}{%
  \raise 4pt \hbox {.}
  \mkern 6mu
  \raise 1pt \hbox {.}
  \mkern 6mu
  \raise -2pt \hbox {.}
}
\makeglossaries

\setkeys{Gin}{width=1.0\columnwidth}

\makeatother

\usepackage{babel}
  \providecommand{\definitionname}{Definition}
  \providecommand{\lemmaname}{Lemma}
  \providecommand{\propositionname}{Proposition}
  \providecommand{\remarkname}{Remark}
\providecommand{\theoremname}{Theorem}
\providecommand{\conjecturename}{Conjecture}
\providecommand{\assumptionname}{Assumption}
\providecommand{\algorithmname}{Algorithm}

\begin{document}

 \title{Centralized active tracking of a Markov chain with unknown dynamics
 \thanks{Mrigank Raman and Ojal Kumar are with the Mathematics department, and Arpan Chattopadhyay is associated with the EE department and the  Bharti School of Telecom  Technology and Management,   IIT  Delhi. Email: \{mt1170736, mt1170741 \}@maths.iitd.ac.in, arpanc@ee.iitd.ac.in. This work  was funded by the PDA and PLN6R budget head, IIT Delhi.}
}

\author{
Mrigank Raman, Ojal Kumar, Arpan Chattopadhyay
\vspace{-5mm}
}

\maketitle

\ifdefined\SINGLECOLUMN
	\setkeys{Gin}{width=0.5\columnwidth}
	\newcommand{\figfontsize}{\footnotesize} 
\else
	\setkeys{Gin}{width=1.0\columnwidth}
	\newcommand{\figfontsize}{\normalsize} 
\fi

\begin{abstract} 
In this paper, selection of an active sensor subset for tracking a discrete time, finite state Markov chain having an unknown transition probability matrix (TPM) is considered. A total of $N$ sensors are available for making observations of the Markov chain, out of which a subset of sensors are activated each time in order to perform reliable estimation of the process. The trade-off is between activating more sensors to gather more observations for the remote estimation, and restricting sensor usage in order to save energy and bandwidth consumption. The problem is formulated as a constrained minimization  problem, where the objective is the long-run averaged mean-squared error (MSE) in estimation, and the constraint is on sensor activation rate. A Lagrangian relaxation of the problem is solved by an artful blending of two tools: Gibbs sampling  for MSE minimization and an on-line version of expectation maximization (EM)  to estimate the unknown TPM. Finally, the Lagrange multiplier is updated using slower timescale stochastic approximation   in order to satisfy the sensor activation rate constraint. The on-line EM algorithm, though adapted from literature, can estimate vector-valued parameters even under  time-varying dimension of the sensor observations. Numerical results demonstrate approximately $1$~dB better error   performance than uniform sensor  sampling and comparable error performance (within $2$~dB bound) against complete sensor observation. This makes the proposed algorithm amenable to practical implementation.
\end{abstract}
\idx{ Active tracking, sensor selection, stochastic approximation, Gibbs sampling, on-line expectation maximization.}

\section{Introduction}\label{section:introduction}
Remote estimation of physical processes via sensor observations is an integral part of cyber-physical systems. These estimates are typically fed to some controller in order to control a physical process or system. Typical applications of remote estimation include object tracking,  environment monitoring, industrial process monitoring and control, state estimation in smart grid, system identification and disaster management. One key challenge is such remote estimation problems is that the sensors are required to perform high-quality sensing, control, communication, and tracking, but they are constrained in terms of energy and bandwidth availability. Hence, it is necessary to activate only the most informative sensor subset at each time, so that a good compromise is achieved between the fidelity of the estimates and energy/bandwidth usage by sensors.

Herein, we consider the problem of designing a {\em low-complexity} algorithm for dynamically activating an optimal sensor subset, that minimizes the time-averaged MSE under a sensor activation rate  constraint reflecting a constraint on the total energy consumed across sensors. The setup is {\em centralized} in the sense that sensors directly report their observations to a remote estimator; in distributed tracking, there can be multiple nodes, each individually estimating a process, via information exchanged over a multi-hop mesh network. In this paper, we consider {\em centralized} tracking of a Markov chain with unknown TPM, and solve the problem via a combination of Gibbs sampling, stochastic approximation, and on-line EM. We also work out the problem for the known TPM case. While these algorithms are motivated by solid theoretical consideration, numerical results demonstrate promising MSE performance despite reduced  complexity.

\begin{figure}
    \centering
    \includegraphics[height=3cm, width=8cm]{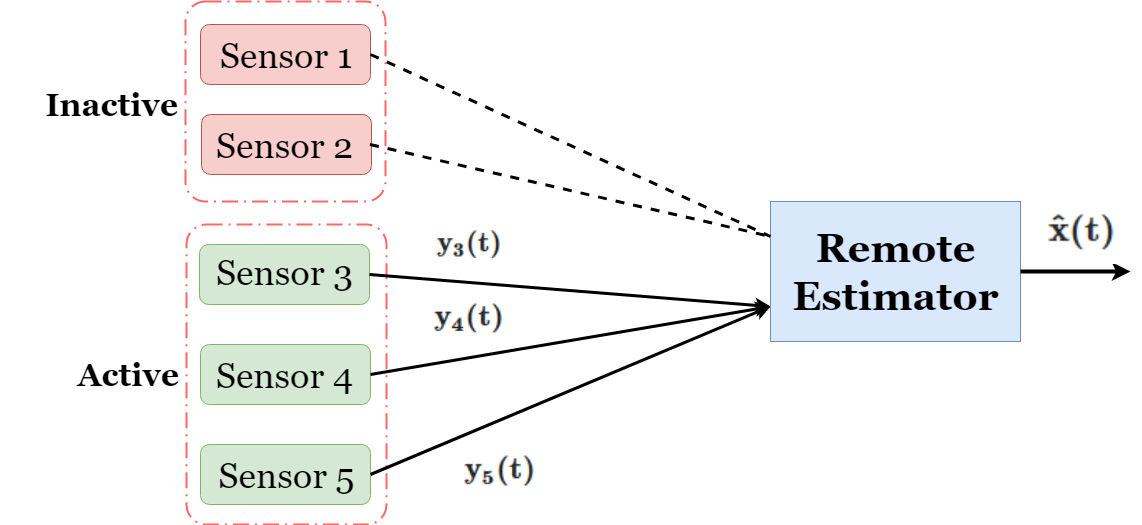}
    \caption{Remote state estimation with active sensing.}
    \label{fig:network-topology-centralized}
    \vspace{-5mm}
\end{figure}

\subsection{Literature survey} 
Active sensor subset selection for process tracking may be either centralized or distributed. In centralized tracking, sensors send observations to a remote estimator which estimates the process. In distributed tracking, a number of  nodes, each having  a number of sensors, constitute a connected, multihop network; each  node   estimates the global state using its local sensor observation and the information coming from the adjacent nodes. 

There have been considerable 
recent work on centralized active tracking of a process with several  applications; see \cite{wang2016efficient} for application on sensor networks,  \cite{schnitzler-etal15sensor-selection-crowdsensing} for mobile crowdsensing, \cite{wang2011information} for body-area sensing application  and \cite{armaghani2011dynamic} for  target tracking application. Even when the process does not have time-variation, calculating the estimation
error given sensor observations   and  finding the
best  subset of sensors poses some serious technical challenges.  To address these challenges, the authors of  \cite{wang2016efficient} provided  a  lower bound on performance, and used a greedy algorithm for subset selection. On the other hand, there have been a number of recent work on centralized active tracking of a time-varying process: \cite{krishnamurthy07structured-threshold-policies} for single sensor selection by a centralized controller  to track a Markov chain, Markov decision process formulation for sensor subset selection in \cite{daphney-etal14active-classification-pomdp} to track a Markov chain with known TPM, energy aware  active sensing  \cite{daphney-etal13energy-efficient-sensor-selection}, existence and structure of active subset selection policy under linear quadratic Gaussian (LQG) model \cite{wu-arapostathis08optimal-sensor-querying}, 
 active sensing for a linear process with unknown  Gaussian noise statistics  via Thompson sampling  \cite{schnitzler-etal15sensor-selection-crowdsensing}, etc. The paper  \cite{gupta-etal06stochastic-sensor-selection-algorithm} considered the model  where an observation is shared across multiple sensors. Recently, there have been a series of work for i.i.d. process tracking:  \cite{chattopadhyay2017optimal}   using Gibbs sampling based subset selection for an i.i.d. process with known distribution, \cite{chattopadhyay2018optimal} for learning an unknown parametric distribution of the process via stochastic approximation (see \cite{borkar08stochastic-approximation-book}), and an extended version \cite{chattopadhyay2020dynamic} of these two papers. On the other hand,  the authors of \cite{Globalsip} have proposed an algorithm for  distributed tracking a Markov chain with {\em known} TPM using tools from stochastic approximation, Gibbs Sampling and Kalman-consensus filter.

\subsection{Our Contribution}
Following points are our contribution in the paper:
\begin{enumerate}
    \item We provide an online learning algorithm for active sensing to track a Markov chain with {\em unknown TPM}.  The unknown TPM is learnt via  online EM (see \cite{cappe2011online}), suitably adapted for variable dimension of observations due to active sensing. Gibbs sampling (see \cite{bremaud2013markov}) is used for {\em low-complexity} sensor activation, and stochastic approximation is used for meeting the sensor activation constraint. 
     \item An interesting trick to handle variable dimension of observations was to maintain a global library of unknown parameter estimates, and update only the relevant ones in an asynchronous fashion, depending on sensor activation and observation sequence. This idea was absent in the online EM algorithm \cite{cappe2011online}.
     \item Another interesting feature of the algorithm which we have proposed is running various iterates in multiple timescales.
\end{enumerate}

\subsection{Organisation}
We have organised our paper in the following manner. In Section~\ref{section:system-model}, we have described system model. The necessary mathematical background is summarised in Section~\ref{section:background}. In Section~\ref{section:GEM-algorithm} we have provided active sensing, state estimation and learning algorithms. Numerical results are provided in Section~\ref{section:numerical-work} and finally we have conclusion in Section~\ref{section:conclusion}.

\section{System Model}\label{section:system-model}
Bold capital and bold small letters will represent matrices and vectors respectively, whereas sets will be represented by calligraphic font throughout this paper.

\subsection{Sensing and observation model}
We consider  a remote estimation setting  as in   Figure~\ref{fig:network-topology-centralized}. The set of sensors is $\mathcal{N}=\{1, 2, 3, . . ., N\}$. The sensors are used to sense a discrete time process $\{\bm{x}(t)\}_{t \geq 0}$, which is  a time-homogeneous positive recurrent Markov chain with $q$ states. For the sake of mathematical convenience, we denote the $j$-th state as $\bm{e_j}$ which is the $j$-th standard basis (column) vector of length $q$, with  $1$ at the $j$-th coordinate and $0$ everywhere else. Hence, the state space of $\{\bm{x}(t)\}_{t \geq 0}$ becomes $\mathcal{X}:=\{\bm{e_j}: j=1,2,\cdots,q\}$. The TPM of $\{\bm{x}(t)\}_{t \geq 0}$ is denoted by $\bm{A}'$ (transpose of $\bm{A}$), which is {\em unknown} to the remote estimator. 

Note that: 
\begin{equation}
    \bm{x}(t+1)=\bm{A}\bm{x}(t)+\underbrace{\bm{x}(t+1)-\bm{A}\bm{x}(t)}_{\doteq \bm{w}(t)}
\end{equation}
where $\bm{w}(t)$ is a zero-mean process noise (non-Gaussian).

At time $t$, let $\bm{b}(t) \in \{0,1\}^N$ denote the activation status of $N$ sensors; if $b_k(t)=1$, then the $k$-th sensor is active, otherwise it is inactive. Let $\bm{b}_{-k}(t) \in \{0,1\}^{N-1}$ be the same as $\bm{b}(t)$, except that $b_k(t)$ is removed. Also, let $(\bm{b}_{-k}(t),0)$ be the same as $\bm{b}(t)$ except that $b_k(t)=0$, and let us assume similar notation for $(\bm{b}_{-k}(t),1)$.  An active sensor makes an observation and communicates that observation to the remote estimator, whereas an inactive sensor does neither of these. If $b_k(t)=1$  and if $\bm{x}(t)=\bm{e_i}$, then the observation $\bm{y}_k(t)$ from sensor~$k$ follows a Gaussian distribution as  $\bm{y}_k(t) \sim \mathcal{N}(\bm{\mu}_{k,i}, \bm{Q}_{k,i}) \in \mathbb{R}^{n_k \times 1}$. Mathematically, we can write the observation coming from sensor~$i$ at time~$t$ as:
\begin{eqnarray}
\bm{y}_k(t)=
\begin{cases}
\bm{H}_k \bm{x}(t)+\underbrace{\bm{v}_k(t)}_{\sim \mathcal{N}(\bm{0}, \bm{Q}_{k,i})},& b_k(t)=1, \bm{x}(t)=\bm{e_i} \\
\emptyset,&  b_k(t)=0\\
\end{cases} \label{eqn:observation-equation}
\end{eqnarray}
where $\bm{H}_k \doteq [\bm{\mu}_{k,1}: \bm{\mu}_{k,2}: \cdots : \bm{\mu}_{k,q}] \in \mathbb{R}^{n_k \times q}$ is the observation matrix of sensor~$k$, and  $\bm{v}_k(t)$ denotes the Gaussian observation noise at sensor~$k$. We assume that $\bm{v}_k(t)$ is independent across sensors and time. {\em We will consider the cases where  $(\bm{\mu}_{k,i}, \bm{Q}_{k,i})$ can be either known or unknown.}

The collection of observations $ \{\bm{y}_k(t):b_k(t)=1, k \in \mathcal{N} \}$, arranged as a column vector via vertical concatenation,  is called $\bm{y}(t)$. This $\bm{y}(t)$ is collected by the remote estimator to estimate $\hat{\bm{x}}(t)$ at each time $t$. {\em The estimate $\hat{\bm{x}}(t)$ can be viewed as a belief   vector on $\mathcal{X}$.}

\subsection{The optimization problem}
Let $\tilde{\pi}_1$ and $\tilde{\pi}_2$ denote two generic rules (deterministic or randomized) for sensor activation and process estimation, respectively. In this paper, we seek to solve the following problem:

\begin{eqnarray}\label{eqn:constrained-optimization}
&& \min_{\tilde{\pi}_1,\tilde{\pi}_2} \limsup_{T \rightarrow \infty} \frac{1}{T}\sum_{t=0}^{T-1}\mathbb{E}||\hat{\bm{x}}(t)-\bm{x}(t)||^2 \nonumber\\
&\text{ such that }& \limsup_{T \rightarrow \infty} \frac{1}{T}\sum_{t=0}^{T-1} \mathbb{E} ||\bm{b}(t)||_1 \leq \bar{N}
\end{eqnarray}

By standard Lagrange multiplier theory, problem~\eqref{eqn:constrained-optimization} can be solved by solving the following relaxed problem:
\begin{eqnarray}\label{eqn:unconstrained-optimization}
&& \min_{\tilde{\pi}_1,\tilde{\pi}_2} \limsup_{T \rightarrow \infty} \frac{1}{T}\sum_{t=0}^{T-1}\mathbb{E}||(\hat{\bm{x}}(t)-\bm{x}(t)||^2 + \lambda ||\bm{b}(t)||_1 )
\end{eqnarray}
under a suitable Lagrange multiplier $\lambda^*$ such that the inequality constraint is met with equality.

\section{Background}\label{section:background}
In this section, we provide a basic background that will be useful in solving \eqref{eqn:unconstrained-optimization}. 

{\bf Gibbs sampling:} Let us assume (for the sake of illustration) that $\bm{x}(t)$ is i.i.d. with known distribution. Then, there exists an optimal   $\bm{b}^*$ such that, using  $\bm{b}(t)=\bm{b}^*$ over the entire time horizon along with MMSE estimation is optimal for \eqref{eqn:unconstrained-optimization}. Thus, the problem reduces to: 
\begin{eqnarray}
\min_{\bm{b} \in \{0,1\}^N} \underbrace{f(\bm{b})+\lambda ||\bm{b}||_1}_{\doteq h(\bm{b})}
\end{eqnarray}
where $f(\bm{b})$ is the MSE under sensor activation vector $\bm{b}$, and $h(\bm{b})$ is the cost under this activation vector. In order to avoid searching over $2^N$ possible activation vectors, the authors of \cite{chattopadhyay2017optimal} used Gibbs sampling for sensor activation. Gibbs sampling generates a Markov chain $\{\bm{b}(t)\}_{t \geq 0}$ whose stationary  distribution is $\pi_{\beta}(\bm{b})\doteq \frac{e^{-\beta h(\bm{b})}}{\sum_{\tilde{\bm{b}} \in \{0,1\}^N} e^{-\beta h(\tilde{\bm{b}})}}$ with a parameter $\beta>0$ interpreted as the inverse temperature in statistical physics. Note that, $\lim_{\beta \rightarrow \infty}\pi_{\beta}(\bm{b}^*)=1$ if the unique minimizer for $h(\cdot)$ is  $\bm{b}^*$ . Hence, for sufficiently large $\beta$, Gibbs sampling under steady state selects  $\bm{b}^*$ with high probability, and we obtain a near-optimal solution of \eqref{eqn:unconstrained-optimization}. At any time $t$, Gibbs sampling randomly selects sensor~$k \in \mathcal{N}$ with uniform distribution, and sets $b_k(t)=1$ with probability $\frac{e^{-\beta h(\bm{b}_{-k},1)}}{e^{-\beta h(\bm{b}_{-k},1)}+e^{-\beta h(\bm{b}_{-k},0)}}$ and $b_k(t)=0$ otherwise. Then the $k$-th sensor is activated accordingly, and the activation status of other sensors remain unchanged. Finally, the constrained problem~\eqref{eqn:constrained-optimization} was solved by using a stochastic approximation (see \cite{borkar08stochastic-approximation-book}) iteration $\lambda(t+1)=\lambda(t)+a(t)(||\bm{b}(t)||_1-\bar{N})$, to satisfy the activation constraint.

{\bf Expectation maximization:} The expectation maximization (EM) algorithm (see \cite{hajek-lecture-note}) is used to estimate an unknown parameter $\bm{\theta}$ from noisy observation $\bm{y}$ of a random vector $\bm{x}$ having a parametric distribution with unknown parameter $\bm{\theta}$. It maintains an iterate $\bm{\theta}(t)$ at iteration $t$. In the E step, $\mathbb{E}(\log p(\bm{x}|\bm{\theta})| \bm{y}, \bm{\theta}(t))$ is computed, and this is maximized over $\bm{\theta}$ to obtain $\bm{\theta}(t+1)$ in the M step. It was shown in \cite{hajek-lecture-note} that, under certain regularity conditions, $\bm{\theta}(t)$ converges to the set of stationary points such that $\pdv{p(\bm{y})}{\bm{\theta}}=\bm{0}$. Later, the authors of \cite{cappe2011online} proposed one online EM algorithm for hidden Markov model; their model is similar to our process and observation models, except that they do not consider active sensing and  consider {\em scalar} observations of fixed dimension. However, due to active sensing, our problem allows variable dimension of observations, which requires some nontrivial modification of the algorithm of \cite{cappe2011online}. 
\section{The GEM algorithm}\label{section:GEM-algorithm}
In this section, we propose an algorithm called GEM (Gibbs Expectation Maximization) to solve \eqref{eqn:constrained-optimization}. Since the algorithm is technically involved, we will first describe the major components and concepts related to the algorithm and finally provide a summarised version of the complete algorithm.

\subsection{Key components of the algorithm}

\subsubsection{Some useful notation} 
\begin{center}
    \begin{tabular}{||m{2.5cm}|m{5.4cm}||}
    \hline
    \begin{center}
        Symbols
    \end{center}{}
     &
    \begin{center}
         Meaning
    \end{center}  \\
    \hline
    $\bm{b}(t)$  & Activation status of $N$ sensors \\
    \hline
    $\lambda(t)$  & Estimate of the cost of a sensor activation \\
    \hline
    $\hat{\bm{A}}_t$ &Running Estimate of Transpose of Transition Probability Matrix \\
    \hline
    $h^{(t)}(\bm{b}), f^{(t)}(\bm{b})$ & Cost and MSE estimates under activation vector $\bm{b}$\\
    \hline
    $\bm{\hat{\mu}}_{k,i}(t), \bm{\hat{Q}}_{k,i}(t)$ & Running Estimate of mean and covariance of $\bm{y}_{k}(t)$  where $\bm{x(t)} = \bm{e}_i$ \\
    \hline
    $\bm{\hat{\Lambda}}_t$ & Running estimate of   the observation matrix $[\bm{H}_1' : \bm{H}_2': \cdots: \bm{H}_N' ]$ \\
    \hline
    $\bm{\hat{\Psi}}_i(t)$ & $blkdiag(\hat{\bm{Q}}_{1,i}(t), \hat{\bm{Q}}_{2,i}(t), \cdots, \hat{\bm{Q}}_{N,i}(t))$ \\
    \hline
    $\bm{\hat{M}^{\bm{b}(t)}}_i(t), \bm{\hat{\Psi}^{\bm{b}(t)}}_i(t)$ & Running estimates of those components of $\bm{\hat{\Lambda}}_t$ and  $\bm{\hat{\Psi}}_i(t)$, that correspond to the active sensors under activation vector $\bm{b(t)}$ \\
    \hline
    \end{tabular}

\end{center}

The proposed algorithm maintains running estimates $\hat{\bm{\mu}}_{k,i}(t)$ and $\hat{\bm{Q}}_{k,i}(t)$ for $\bm{\mu}_{k,i}$ and $\bm{Q}_{k,i}$, respectively. 
Equivalently, it maintains an estimate 
$\hat{\bm{\Lambda}}_t$ of the matrix $[\bm{H}_1' : \bm{H}_2': \cdots: \bm{H}_N' ]$ where $\bm{H}_k \doteq [\bm{\mu}_{k,1}: \bm{\mu}_{k,2}: \cdots : \bm{\mu}_{k,q}] $. The algorithm also maintains $q$ block-diagonal matrices $\{\hat{\bm{\Psi}}_i (t): 1 \leq i \leq q\}$ where 
$\hat{\bm{\Psi}}_i(t) \doteq blkdiag(\hat{\bm{Q}}_{1,i}(t), \hat{\bm{Q}}_{2,i}(t), \cdots, \hat{\bm{Q}}_{N,i}(t))$ (block diagonal matrix consisting of these $N$ covariance matrix estimates at time $t$).
 For an activation vector $\bm{b}$, we also define another matrix  $\hat{\bm{\Psi}}_i^{\bm{b}}(t) \doteq blkdiag\{\hat{\bm{Q}}_{k,i}(t): 1 \leq k \leq N, b_k=1\}$  which can be  extracted from $\hat{\bm{\Psi}}_i(t)$. Similarly, we define $\hat{\bm{\mu}}_i^{\bm{b}}(t)$ as an estimate of  of the column vector $vertcat(\hat{\bm{\mu}}_{k,i}: 1 \leq k \leq N, b_k=1)$ (vertical concatenation of these column vectors). 
Clearly, given $\bm{x}(t)=\bm{e}_i$ and activation vector $\bm{b}(t)$, 
our algorithm assumes that $\bm{y}(t) \sim \mathcal{N}(\hat{\bm{\mu}}_i^{\bm{b}(t)}(t), \hat{\bm{\Psi}}_i^{\bm{b}(t)}(t))$. For a given activation vector b we also define the matrix $\hat{\bm{M}}^{\bm{b}}(t) \doteq [\hat{\bm{\mu}}_{1}^{\bm{b}}(t):\hat{\bm{\mu}}_{2}^{\bm{b}}(t):  \cdots:\hat{\bm{\mu}}_{q}^{\bm{b}}(t) ]$. The algorithm also maintains an estimate $\hat{\bm{A}}_t$ for $\bm{A}$.

The algorithm also maintains the iterates $h^{(t)}(\bm{b}) \forall \bm{b} \in \{0,1\}^N$, $f^{(t)}(\bm{b}) \forall \bm{b} \in \{0,1\}^N$, and $\lambda(t)$ (see Section~\ref{section:background}), as estimates of $h(\bm{b}) \forall \bm{b} \in \{0,1\}^N$, $f(\bm{b}) \forall \bm{b} \in \{0,1\}^N$, and $\lambda^*$, respectively. Estimate of the MSE under activation vector $\bm{b}$ can be denoted by  $f^{(t)}(\bm{b}) \forall \bm{b} \in \{0,1\}^N$ at any time~$t$.  The quantities $h^{(t)}(\bm{b}) \forall \bm{b} \in \{0,1\}^N$ are used as cost in Gibbs sampling at time~$t$ to decide the sensor activation set. The Lagrange multiplier $\lambda(t)$ is updated using stochastic approximation so that the activation constraint in \eqref{eqn:constrained-optimization} satisfies the equality constraint.

We define $\nu_{\bm{b}}(t) \doteq \sum_{\tau=1}^t \mathbb{I} \{\bm{b}(\tau)=\bm{b}\}$ as the number of times the activation vector $\bm{b}$  is used up to time $t$. 

\subsubsection{Step sizes}\label{subsubsection:step-size}
The algorithm maintains two non-increasing, positive step size sequences $\{\alpha(t): t=0,1,2, \cdots\}$ and $\{\gamma(t): t=0,1,2, \cdots\}$ for running multi-timescale stochastic approximation updates. The step size sequences satisfy the following properties: (i) $\sum_{t=0}^{\infty}s(t)=\infty, s \in \{\alpha, \gamma\}$, (ii) $\sum_{t=0}^{\infty}s^2(t)<\infty, s \in \{\alpha, \gamma\}$, (iii) $\lim_{t \rightarrow \infty}\frac{\gamma(t)}{\alpha(t)}=0$. The first two requirements are standard for stochastic approximation. The third condition is required for timescale separation; the $f^{(t)}(\cdot)$ update uses step-size $\alpha(t)$, and the $\lambda(t)$ update and online EM updates will use step size $\gamma(t)$.

\subsubsection{Gibbs sampling}\label{subsubsection:gibbs-sampling}
At time $t$, pick a random sensor $j_t \in \mathcal{N}$ uniformly and independently. For sensor $j_t$, choose $\bm{b}_{j_{t}}(t)$ = 1 with probability  $p_t=\frac{e^{-\beta h^{(t)}(\bm{b}_{-j_t}(t-1),1)}}{e^{-\beta h^{(t)}(\bm{b}_{-j_t}(t-1),1)} + e^{-\beta h^{(t)}(\bm{b}_{-j_t}(t-1),0)}}$ and choose $b_{j_{t}}(t)=0$   with probability $(1-p_t)$. For all $k \neq j_t$, we choose $b_k(t)=b_k(t-1)$. Activate sensors according to $\bm{b}(t)$ and obtain the observations $\bm{y}(t)$.

\subsubsection{$\lambda(t)$ update}\label{subsubsection:Lagrange-multiplier-update}
The Lagrange multiplier is updated as:
\begin{equation}\label{eqn:lambda-update}
    \lambda(t+1)=[\lambda(t)+\gamma(t) (||\bm{b}(t)||_1 -\bar{N})]_0^l
\end{equation}
The iterates are projected onto an interval which is compact namely $[0,l]$ to ensure boundedness, where $l>0$ is a sufficiently large number. The intuition behind this approach is that, if $||\bm{b}(t)||_1 >\bar{N}$, then $\lambda(t)$ (the cost of a sensor activation) is increased, and $\lambda(t)$ is decreased otherwise.

\subsubsection{State estimation}\label{subsubsection:state-estimation} 
We use a {\em Kalman-like state estimator} from \cite{daphney-etal14active-classification-pomdp}, designed to track a Markov chain. In this algorithm, $\hat{\bm{x}}_{t+1|t}$ and $\hat{\bm{x}}_{t|t}$ denote the estimates of $\bm{x}(t+1)$ and $\bm{x}(t)$ respectively, given $\{\bm{y}(0),\bm{y}(1),\cdots,\bm{y}(t)\}$; here $\hat{\bm{x}}_{t|t}$ is basically the final estimate $\hat{\bm{x}}(t)$ declared by the estimator at time~$t$, and $\hat{\bm{x}}_{t|t-1}$ is an intermediate estimate at time~$t$. Additionally, 
$\bm{\Sigma}_{t|t}$ and $\bm{\Sigma}_{t|t-1}$ will denote covariance matrices estimates of the estimation and prediction errors $(\hat{\bm{x}}_{t|t}-\bm{x}(t))$ and $(\hat{\bm{x}}_{t|t-1}-\bm{x}(t))$ respectively. Unlike standard Kalman filter where the observation dimension is fixed, this Kalman-like estimator has variable observation dimension depending on $\bm{b}(t)$, and the gain and error covariance matrix updates also take into account $\bm{b}(t)$ at time $t$.

\vspace{2mm}
\hrule
\hrule
\vspace{1mm}
\noindent{\bf State Estimation algorithm}
  \vspace{1mm}
 \hrule
 \hrule
  \vspace{2mm}
  \noindent{\bf Recursion:} For each $t \geq 0$, do:
  \begin{enumerate}
      \item $\hat{\bm{{x}}}_{t|t-1}=\bm{A}\hat{\bm{x}}_{t-1|t-1}$,
      
      \noindent \footnotesize{(Comment: State estimate at time $t$, keeping observations up to time $(t-1)$ in account.)}
      \normalsize
      \item $\textbf{y}_{t|t-1}=\bm{\hat{M}}^{\bm{b}(t)}(t)\hat{\bm{x}}_{t|t-1}$,
      \item $\bm{\Sigma}_{t|t-1}= diag(\hat{\bm{x}}_{t|t-1})-\hat{\bm{x}}_{t|t-1}\hat{\bm{x}}_{t|t-1}'$,
      
      \noindent \footnotesize{(Comment: Estimate of error covariance matrix at time $t$, keeping observations up to time $(t-1)$ in account.) }
      
      \normalsize
      \item $\tilde{\bm{\Psi}}_{t} = \sum_{i=1}^{q} \hat{\bm{x}}_{t|t-1}(i) \bm\hat{{\Psi}}_i^{\bm{b}(t)}$,
      
      \noindent \footnotesize{(Comment: Estimate of the observation noise covariance matrix under activation vector $\bm{b}(t)$, averaged over the belief $\hat{\bm{{x}}}_{t|t-1}(.)$ on the state of the process.)}
      \normalsize
      \item $\textbf{G}_{t}=\bm{\Sigma}_{t|t-1}\bm{\hat{M}}^{\bm{{b}(t)}}(t))^{'}*(\bm{\hat{M}}^{\bm{{b}(t)}}(t))\bm{\Sigma}_{t|t-1}((\bm{\hat{M}}^{\bm{{b}(t)}}(t))^{'}+\tilde{\bm{\Psi}}_{t})^{-1}$,
      
      \noindent \footnotesize{(Comment: Kalman gain update.)}
      \normalsize
      \item Compute $\hat{\bm{x}}_{t|t}=\hat{\bm{x}}_{t|t-1}+\textbf{G}_t(\bm{y}(t)-\bm{y}_{t|t-1}), $ and project it on the probability simplex.
      
      \noindent \footnotesize{(Comment: State estimate at time $t$, keeping observations up to time $(t-1)$ in account. Projection ensures that the estimate is a valid probability belief vector on the state space.)}
      \normalsize
      
      \item $\bm{\Sigma}_{t|t}=diag(\hat{\bm{x}}_{t|t})-\hat{\bm{x}}_{t|t}\hat{\bm{x}}_{t|t}'$.
      
      \noindent \footnotesize{(Comment: Estimate of error covariance matrix at time $t$, keeping observations up to time $t$ in account.)}
      \normalsize
      
  \end{enumerate}
  \hrule
  \hrule

\subsubsection{$f^{(t)}(\cdot)$ update}\label{subsubsection:f-update}
At time~$t$, MSE estimate $f^{(t)}(\cdot)$ (for sensor subset $\bm{b}(t)$ only) is updated using the following equation: 

\footnotesize
\begin{equation}\label{eqn:f-update}
    f^{(t+1)}(\bm{b})=[f^{(t)}(\bm{b})+\mathbb{I}\{\bm{b}=\bm{b}(t)\} \alpha(\nu_{\bm{b}}(t)) (\mbox{Tr}(\bm{\Sigma}_{t|t})-f^{(t)}(\bm{b}))]_0^l
\end{equation}
\normalsize

\subsubsection{Online EM for parameter estimation}\label{subsubsection:online-EM}
Online EM requires an initial distribution $\pi$ for the Markov chain.   However, unlike \cite{cappe2011online}, here we have {\em vector}-valued observations whose dimension change over time. Also, the unknown parameters  $\hat{\bm{\mu}}_{i}^{\bm{b}(t)}$ and $\hat{\bm{\Psi}}_i^{\bm{b}(t)}$ (known $\hat{\bm{\mu}}_{i}^{\bm{b}(t)}$ and $\hat{\bm{\Psi}}_i^{\bm{b}(t)}$ can be handled in a straightforward way) will also have different dimensions for different values of $t$. This requires an asynchronous update of various components of the unknown parameters, depending on the currently active sensors and their observations. 

Since the online EM algorithm is heavy in notation, we have made some explanatory comments in between the steps of the algorithm. See \cite{cappe2011online} for a detailed understanding.

The algorithm  requires the auxiliary functions $\hat{\phi}, \hat{\rho}^{\bm{A}}, \hat{\rho}^{g}, \hat{{S}}^{\bm{A}}$ and $ \hat{\bm{S}}^{g} $, and $g_t(\bm{e}_{i},\bm{y})$. We also define  $\bm{y}^2(t)=blkdiag\{\bm{y}_{k}(t)\bm{y}_k'(t) : 1 \leq k \leq N, b_k(t)=1\}$. In general, we need $\bm{y}^d(t)$  in the algorithm where $d \in \{0,1,2\}$; we define $\bm{y}^{0}(t) = 1 (scalar), and \bm{y}^{1}(t) \doteq \bm{y}(t)$.

 \vspace{2mm}
 \hrule
 \hrule
 \vspace{1mm}
 \noindent {\bf Online EM algorithm}
  \vspace{1mm}
 \hrule
 \hrule
  \vspace{2mm}
  
\noindent {\bf Input:} Initial distribution $\bm{\pi}$ of $\{\bm{x}(t)\}_{t \geq 0}$, $\bm{y}(t)$.

\noindent {\bf Initialization:} Initialize $\hat{\bm{A}}_0$
$\bm\hat{{\Psi}}_{i}(0)$ for all $1 \leq i \leq q$ and $\bm\hat{{\Lambda}}_{0}$ randomly and compute, for all $1\leq i,j,k \leq q$  and $0 \leq d \leq 2$,\newline

$\hat{\phi}_{0}(k)=\displaystyle\frac{\pi(k)g_{0}(\bm{e}_{k},\bm{y}_{0})}{\sum\limits_{k_1=1}^{q}g_{0}(\bm{e}_{k'},\bm{y}_{0})}$,\newline

$\hat{\rho}_{0}^{\bm{A}}(i,j,k)=0$,\newline

$\hat{\rho}_{0,d}^{g}(i,k)=\delta_{ik} \bm{y}^{d}(0)$

\noindent {\bf Recursion:}  For $t \geq 1$, for all $1\leq i,j,k \leq q$  and $0 \leq d \leq 2$, DO \\

{\em Approx. Filter Update} 

 $\hat{\phi}_{t}(\bm{e}_k)=\displaystyle\frac{\sum\limits_{k_1=1}^{q}\hat{\phi}_{t-1}(\bm{e}_{k_1})\bm\hat{{A}}^{'}_{t-1}(\bm{e}_{k_1},\bm{e}_k)g_{t-1}(\bm{e}_k,\bm{y}(t))}{\sum\limits_{k_1=1}^{q}\sum\limits_{k_2=1}^{q}\hat{\phi}_{t-1}(\bm{e}_{k_1})\bm\hat{{A}}^{'}_{t-1}(\bm{e}_{k_1},\bm{e}_{k_2})g_{t-1}(\bm{e}_{k_2},\bm{y}(t))}  $\\

\noindent where $g_t(\bm{e}_{i},\bm{y}) \doteq exp[-(\bm{y}-\bm{\hat{\mu}}_{i}^{\bm{b}(t)})[\bm{\Psi}_i^{\bm{b}(t)}]^{-1}(\bm{y}-\bm{\hat{\mu}}_{i}^{\bm{b}(t)})^{T}/2]$\\

\noindent \footnotesize{(Comment: Here $\hat{\phi}_t(\cdot)$ is interpreted as an estimate of the steady state probability distribution of the Markov chain given all observations up to time $t$.)}
\normalsize

{\em E-Step}\newline

\begin{enumerate}
\item $\hat{\rho}_{t}^{\bm{A}}(i,j,k)=\gamma(t)\delta_{jk}\hat{r}_{t}(i|j)+(1-\gamma(t))\sum\limits_{k_1=1}^{q}\hat{\rho}_{t-1}^{\bm{A}}(i,j,k_1)\hat{r}_{t}(k_1|k)$ 

\item $\hat{\rho}_{t,d}^{g}(i,k)=\gamma(t)\delta_{ik}\bm{y}^d(t)+(1-\gamma(t))\sum\limits_{k_1=1}^{q}\hat{\rho}_{t-1,d}^{g}(i,k_1)\hat{r}_{t}(k_1|k)$ 

\noindent where $\hat{r}_{t}(i|j)=\dfrac{\hat{\phi}_{t-1}(\bm{e}_{i})\bm\hat{{A}}^{'}_{t-1}(\bm{e}_{i},\bm{e}_{j})}{\sum\limits_{i_1=1}^{q}\hat{\phi}_{t-1}(\bm{e}_{i_1})\bm\hat{{A}}^{'}_{t-1}(\bm{e}_{i_1},\bm{e}_{j})}$.\\
\end{enumerate}

\noindent \footnotesize{(Comment: $\hat{\rho}_{t}^{\bm{A}}(\cdot,\cdot,\cdot)$ ) and $\hat{\rho}_{t,d}^{g}(\cdot,\cdot)$ together constitute the expectation of a sufficient statistic for the expected log-likelihood involved in the E step. The sufficient statistic is updated over time using Bayes' theorem.  Details can be found in \cite{cappe2011online}.)}
\normalsize

{\em M-Step} \\

\begin{enumerate}
\item $\hat{S}_{t}^{\bm{A}}(i,j)=\sum\limits_{k_1=1}^{q}\hat{\rho}_{t}^{\bm{A}}(i,j,k_1)\hat{\phi}_{t}(\bm{e}_{k_1})$  

\item $\bm{\hat{A}^{'}}_{t}(\bm{e}_{i},\bm{e}_{j})=\dfrac{\hat{S}_{t}^{\bm{A}}(i,j)}{\sum\limits_{j_1=1}^{q}\hat{S}_{t}^{\bm{A}}(i,j_1)}$  

\noindent \footnotesize{(Comment: Update for the estimate of the transition probability matrix.)}
\normalsize

\item $\hat{S}_{t,d}^{g}(i)=\sum\limits_{k_1=1}^{q}\hat{\rho}_{t,d}^{g}(i,k_1)\hat{\phi}_{t}(\bm{e}_{k_1})$  

\item $\bm\hat{{\mu}}_{i}^{\bm{b}(t)}=\dfrac{\hat{S}_{t,1}^{g}(i)}{\hat{S}_{t,0}^{g}(i)}$ 

\noindent \footnotesize{(Comment: Update for the estimate of the sensor observation mean for subset $\bm{b}(t).$)}
\normalsize

\item $\bm\hat{{\Psi}}_i^{\bm{b}(t)}=\dfrac{\hat{\bm{S}}_{t,2}^{g}(i)}{\hat{S}_{t,0}^{g}(i)} - \bm\hat{{\mu}}_{i}^{\bm{b}(t)} (\bm\hat{{\mu}}_{i}^{\bm{b}(t)})'$ 

\noindent \footnotesize{(Comment: Update for the estimate of the sensor observation covariance for subset $\bm{b}(t)$.)}
\normalsize

\item For every active sensor~$k$ such that $b_k(t)=1$ and for all $1 \leq i \leq q$, modify the values of $\hat{\bm{\mu}}_{k,i}(t)$ in $\bm{\hat{\Lambda}_{t-1}}$ and $\hat{\bm{Q}}_{k,i}(t)$ in $\bm{\hat{\Psi}}_i(t-1)$ accordingly.
\end{enumerate}

\hrule 
\hrule

\subsection{The complete algorithm} \label{subsection:complete-algorithm}

\subsubsection{GEM algorithm} 
The complete GEM algorithm is outlined below.

 \hrule
 \hrule
 \vspace{1mm}
 \noindent {\bf The Complete GEM Algorithm}
  \vspace{1mm}
 \hrule
 \hrule

  \noindent{\bf Input:} $\{\alpha(t)\}_{t \geq 0}$, $\{\gamma(t)\}_{t \geq 0}$, $\bm{\pi}$, $\bar{N}$, $\beta$  
  
  \noindent{\bf Initialization:} Initialise $\bm{b(0)}$,  $\lambda(0)\geq{0}$, $\hat{\bm{A}}_0$, $\hat{\bm{\mu}}_{k,i}(0)$ and $\hat{\bm{Q}}_{k,i}(0)$ for all  $1 \leq k \leq N, 1 \leq i \leq q$, 
  
  $\hat{\bm{x}}_{0|-1} = \hat{\bm{x}}_{0|0} = \bm{\pi}$.
  
\noindent {\bf Recursion:}  
For all $t \geq 0$, DO:\\
\begin{enumerate}
\item Use Gibbs Sampling described in Section~\ref{subsubsection:gibbs-sampling} to attain $\bm{b}(t)$, activate sensors accordingly, and collect the corresponding observations $\bm{y}(t)$. 
\item Update $\lambda(t)$ as in Section~\ref{subsubsection:Lagrange-multiplier-update}. 
\item Compute $\hat{\bm{x}}(t)=\hat{\bm{x}}_{t|t}$ as in Section~\ref{subsubsection:state-estimation}, by running the Kalman-like algorithm. 
\item Calculate $f^{(t)}(\bm{b}(t))$ as in Section~\ref{subsubsection:f-update}.
\item Calculate $h^{(t+1)}(\bm{b})=f^{(t+1)}(\bm{b})+\lambda(t+1) ||\bm{b}||_1$
\item Compute the estimates $\hat{\bm{A}}_t$, $\hat{\bm{\mu}}_{k,i}(t)$ and $\hat{\bm{Q}}_{k,i}(t)$ for all  $1 \leq k \leq N, 1 \leq i \leq q$, by using online EM as in Section~\ref{subsubsection:online-EM}.
\end{enumerate}
\hrule
\hrule

\subsubsection{Discussion}
\begin{itemize}
    \item The GEM algorithm runs in multiple timescales (see \cite{borkar08stochastic-approximation-book}).  Gibbs sampling corresponds to the fastest timescale, and the $\lambda(t)$ update and  online EM run in the slowest timescale. Timescale separation is ensured by $\lim_{t \rightarrow \infty}\frac{\gamma(t)}{\alpha(t)}=0$.
    \item $f^{(t)}(\cdot)$ update involves asynchronous stochastic approximation for various sensor subsets. 
    \item Projection on  $[0,l]$ is done to make sure that all iterates remain bounded.
\end{itemize}

\subsubsection{Computational complexity of GEM for every time~$t$} 
At each time, Gibbs sampling and $\lambda(t)$ update require $\mathcal{O}(1)$ computations. Updating $f^{(t)}(\cdot)$ requires $\mathcal{O}(q)$ computations for trace calculation. The approximate filter update step for all states will require $\mathcal{O}(q N^3 \max_{1 \leq k \leq N}\{n_k^3\})$ computations due to the matrix inversion involved in $g_t(\cdot,\cdot)$ calculation. The first step in E step requires $\mathcal{O}(q^4)$ computations. Computational complexity of other steps of online EM as well as state estimation is dominated by these two steps and hence the computational complexity of online EM at each time is $\mathcal{O}(q N^3 \max_{1 \leq k \leq N}\{n_k^3\}+q^4)$, where $n_k$ is the dimension of $\bm{y}_k(t)$ as defined before. The function $h^{(t)}(\cdot)$ needs to be computed only for three vectors as required by Gibbs sampling, and hence this requires $\mathcal{O}(1)$ computations.

\section{Numerical results}\label{section:numerical-work}
We consider number of sensors  $N=20$,  number of states $q=10$, activation constraint $\bar{N}=5$, inverse temperature $\beta=10$, $\alpha(t)=\dfrac{1}{t^{0.7}}$,  $\gamma(t)=\dfrac{1}{t^{0.8}}$ and  $\lambda(0) = 0.1$. The TPM $\bm{A}'$ is chosen randomly  and then the rows are normalized to obtain a stochastic matrix.  The quantities $\{\bm{\mu}_{k,i}, \bm{Q}_{k,i}\}_{1 \leq k \leq N, 1 \leq i \leq q}$ are also chosen randomly. We have not considered larger value of $N$ because it is unlikely (even in current cyber-physical systems) that so many sensors  estimate a single physical process, especially when the observation from each sensor is a vector.

Under this setting, we compare the performance of the following six   algorithms: 
\begin{enumerate}
    \item GEM-K: Here the observation mean and covariances are known (i.e., K), but TPM is unknown.
    \item GEM-UK: Here observation covariances are known, but observation mean  and TPM are  unknown (i.e., UK).
    \item GEM-FO: This is GEM Algorithm with full observation. Here all the sensors are always active. The observation mean and covariances are known, but the TPM is unknown.
    \item GEM-U: This is GEM Algorithm with uniform random sampling of sensors:  at each time $t$, a sensor is activated independently  with probability $\dfrac{\bar{N}}{N}$. The observation mean and covariances are  known, but the TPM is unknown.
    \item GEM-FI: This is GEM with full information of the TPM, observation mean and covariances.
    \item GEN: Here GEN stands for genie. At time $t$, the estimator perfectly knows $\bm{x}(t-1)$, but no observation is available from sensors. In this case, the MSE will be the limiting variance of  $\bm{x}(t)$ given $\bm{x}(t-1)$.
\end{enumerate}

\subsection{Convergence of the algorithms}
\begin{figure}
    \centering
    \includegraphics[height=7cm, width=9cm]{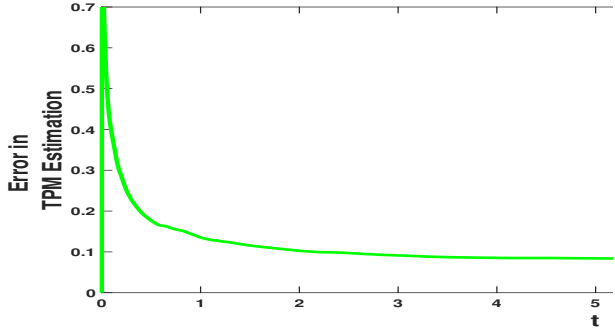}
    \caption{Variation of the TPM estimation error $\frac{1}{t}\sum_{\tau=1}^t ||\bm{A}-\hat{\bm{A}}_{\tau}||_F$ with $t$ for GEM-K.}
    \label{fig:convergence-of-TPM}
\end{figure}

\begin{figure}
    \centering
    \includegraphics[height=7cm, width=9cm]{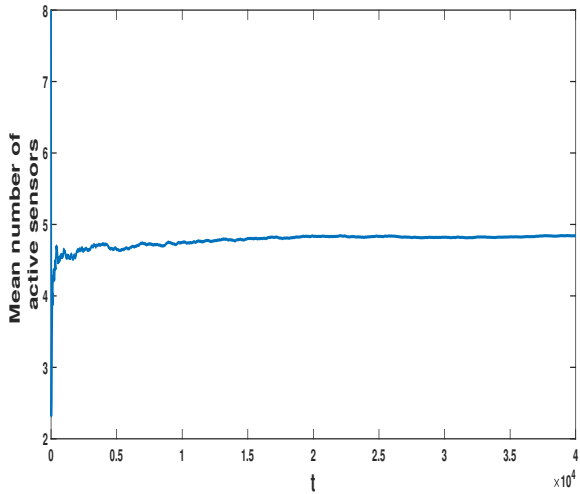}
    \caption{Variation of  $\frac{1}{t}\sum_{\tau=1}^t ||\bm{b}(\tau)||_1$ with $t$ for GEM-K.}
    \label{fig:convergence-of-number-of-active-sensors}
\end{figure}

\begin{figure}
    \centering
    \includegraphics[height=7cm, width=9cm]{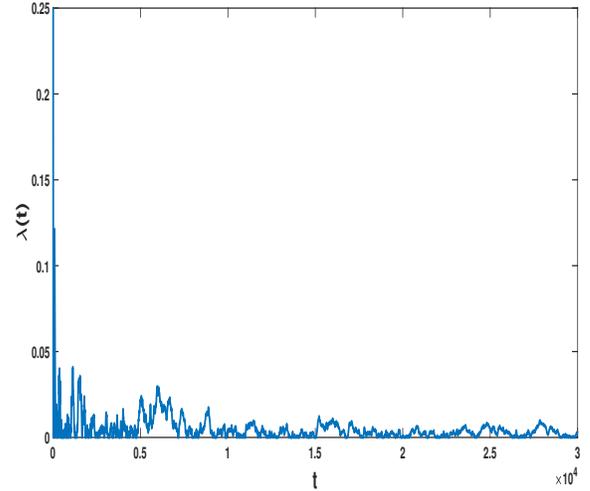}
    \caption{Variation of $\lambda(t)$ with $t$ for GEM-K.}
    \label{fig:convergence-of-lambda}
\end{figure}

Figure~\ref{fig:convergence-of-TPM} shows the convergence of $\hat{\bm{A}}_t$ to $\bm{A}$ for GEM-K. Similarly, we have noticed that $\hat{\bm{A}}_t$ converges to $\bm{A}$ for GEM-FO and GEM-U. However, the TPM estimate does not converge to the true TPM for GEM-UK; instead, it converges to some local optimum as guaranteed by the EM algorithm. For all relevant algorithms, we have noticed that the mean number of active sensors, calculated as $\frac{1}{t}\sum_{\tau=1}^t ||\bm{b}(\tau)||_1$, converges to $\bar{N}$; this has been illustrated only for GEM-K algorithm in Figure~\ref{fig:convergence-of-number-of-active-sensors} and the corresponding $\lambda(t)$ variation is shown in Figure~\ref{fig:convergence-of-lambda}. We observe that $\lambda(t)$ converges at a slower rate.

\begin{figure}
 \centering
 \includegraphics[height=7cm, width=9cm]{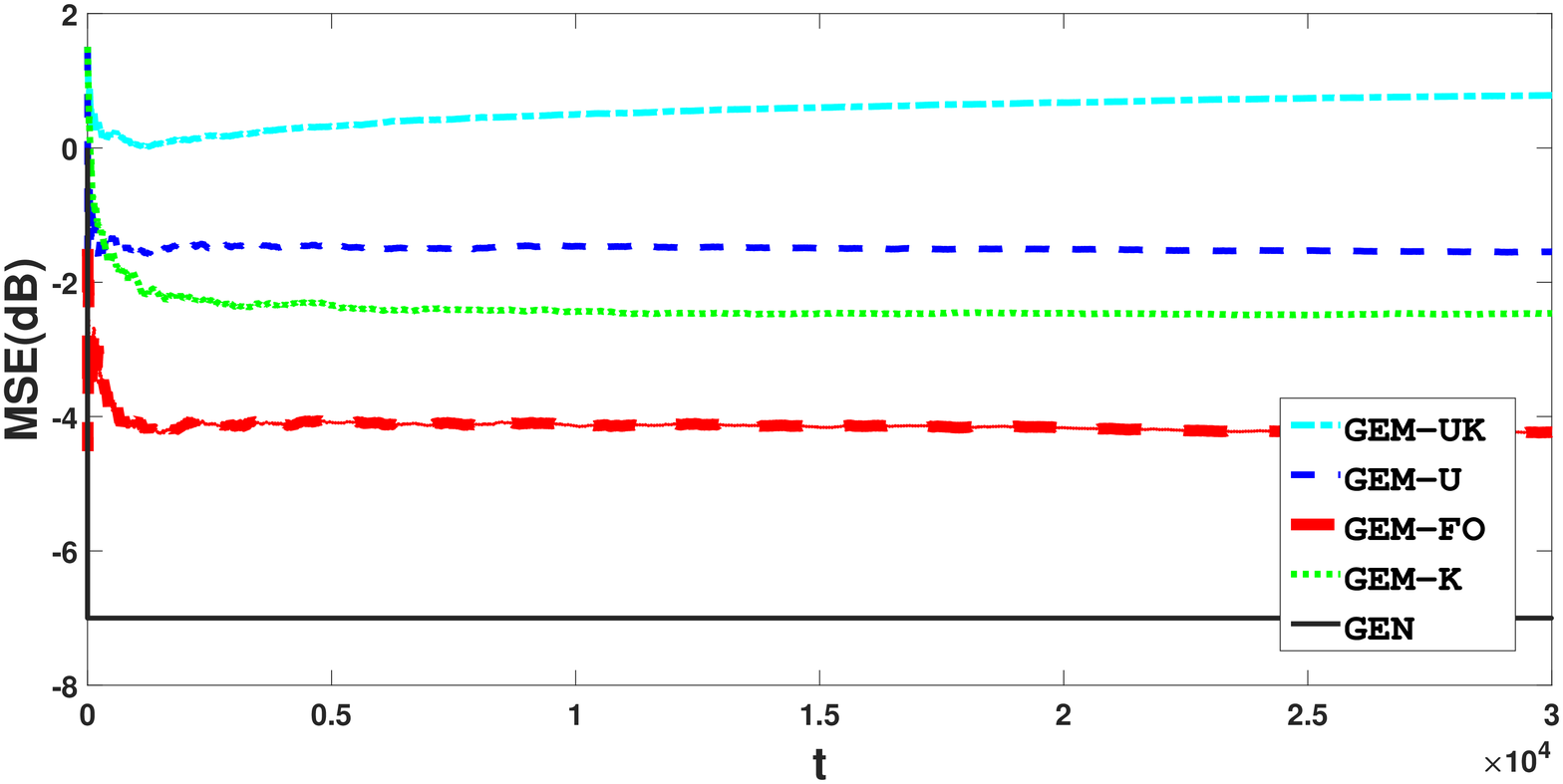}
 \caption{MSE performance comparison among various algorithms.}
 \label{fig:MSE-comparison}
\end{figure}

\subsection{Performance comparison}
In Figure~\ref{fig:MSE-comparison}, we have compared the MSE performance of various algorithms. We observe that, the TPM estimate in GEM-K converges to the true TPM, and hence the asymptotic MSE performance of GEM-K and GEM-FI are same. Hence, we do not show the MSE performance of GEM-FI separately. Figure~\ref{fig:MSE-comparison} shows that, GEN has the best MSE performance due to perfect knowledge of the previous state, and GEM-UK has the worst MSE performance because it converges to a local optimum. GEM-FO performs better than GEM-K because it uses more sensor observation, but it cannot outperform GEN. GEM-U outperforms GEM-UK, since it has knowledge of observation mean and covariances. However, we note in multiple simulations that, on a fair comparison with GEM-K, GEM-U performs worse by approximately $1$~dB; this shows the power of Gibbs sampling against uniform  sampling. 

We  have repeated this for $10$ different instances, and found that the ordering of MSE performances across algorithms remained  unchanged, though the relative performance gaps among the  algorithms varied. However, performance gap of GEM-K was observed to be within $2$~dB of the MSE of GEM-FO, and within several dB from the MSE of GEN. This shows that GEM is very useful for tracking a Markov chain. 

\section{Conclusions}\label{section:conclusion}
We have provided a low-complexity active sensor   selection algorithm for centralized tracking of a  Markov chain with unknown transition probability matrix. The algorithm uses  Gibbs sampling, multi-timescale stochastic approximation, online EM and Kalman-like state estimation to achieve a good compromise among computational complexity, fidelity of estimate, and  energy and bandwidth usage in state estimation. Performance of the algorithm has been validated numerically. We seek to prove convergence of the proposed algorithm, and also extend this work for distributed tracking problems in our future research.

{\small
\bibliographystyle{IEEEtran}
\bibliography{arpan-techreport.bib,erik.bib}

}

\end{document}